\documentclass[conference]{IEEEtran}
\IEEEoverridecommandlockouts
\pdfoutput=1
\usepackage{cite}
\usepackage{amsmath,amssymb,amsfonts}
\usepackage{algorithmic}
\usepackage{graphicx}
\usepackage{array}
\usepackage{textcomp}
\usepackage{booktabs}
\usepackage{multirow}
\usepackage{bigstrut}
\usepackage{xcolor}
\usepackage[numbers]{natbib}
\usepackage[misc]{ifsym}
\def\BibTeX{{\rm B\kern-.05em{\sc i\kern-.025em b}\kern-.08em
    T\kern-.1667em\lower.7ex\hbox{E}\kern-.125emX}}
\begin{document}

\title{AnANet: Modeling Association and Alignment for Cross-modal Correlation Classification}

\author{\IEEEauthorblockN{\Letter Nan Xu\textsuperscript{1,2,3}, Junyan Wang\textsuperscript{3}, Yuan Tian\textsuperscript{1,2}, Ruike Zhang\textsuperscript{1,2}, and Wenji Mao\textsuperscript{1,2}}
\IEEEauthorblockA{\textit{Institute of Automation, Chinese Academy of Sciences}\\
\textit{School of Artificial Intelligence, University of Chinese Academy of Sciences}\\
\textit{Beijing Wenge Technology Co.,Ltd}\\
Beijing 100190, China}
}

\maketitle

\begin{abstract}
	The explosive increase of multimodal data makes a great demand in many cross-modal applications that follow the strict prior related assumption. Thus researchers study the definition of cross-modal correlation category and construct various classification systems and predictive models. However, those systems pay more attention to the fine-grained relevant types of cross-modal correlation, ignoring lots of implicit relevant data which are often divided into irrelevant types. What's worse is that none of previous predictive models manifest the essence of cross-modal correlation according to their definition at the modeling stage. In this paper, we present a comprehensive analysis of the image-text correlation and redefine a new classification system based on implicit association and explicit alignment. To predict the type of image-text correlation, we propose the Association and Alignment Network according to our proposed definition (namely AnANet) which implicitly represents the global discrepancy and commonality between image and text and explicitly captures the cross-modal local relevance. The experimental results on our constructed new image-text correlation dataset show the effectiveness of our model.  
\end{abstract}


\section{Introduction}
With the explosive increase of multimodal data in social media, multimodal data (e.g. image-text pair) is in great demand in many cross-modal applications, such as multimodel event classification \cite{ZEPPELZAUER201645}, multimodel topic labeling \cite{sorodoc2017multimodal}, multimodal sentiment analysis \cite{zadeh2017tensor}, multimodal named entity recognition \cite{zhang2018adaptive},  cross-modal retrieval \cite{zhen2019deep}, multimodal hashtag recommendation \cite{ma2019co}, etc. These works follow the prior assumption that the paired image and text are strictly related and map the multimodal data into the common space via cross-modal correlation learning. But, none of them explicitly define the specific categories of cross-modal correlation during the learning process. Therefore, in practical applications such as social media, these assumptions are fragile and difficult to be strictly satisfied.

\begin{figure}[ht]
	\centering
	\includegraphics[width=1\linewidth]{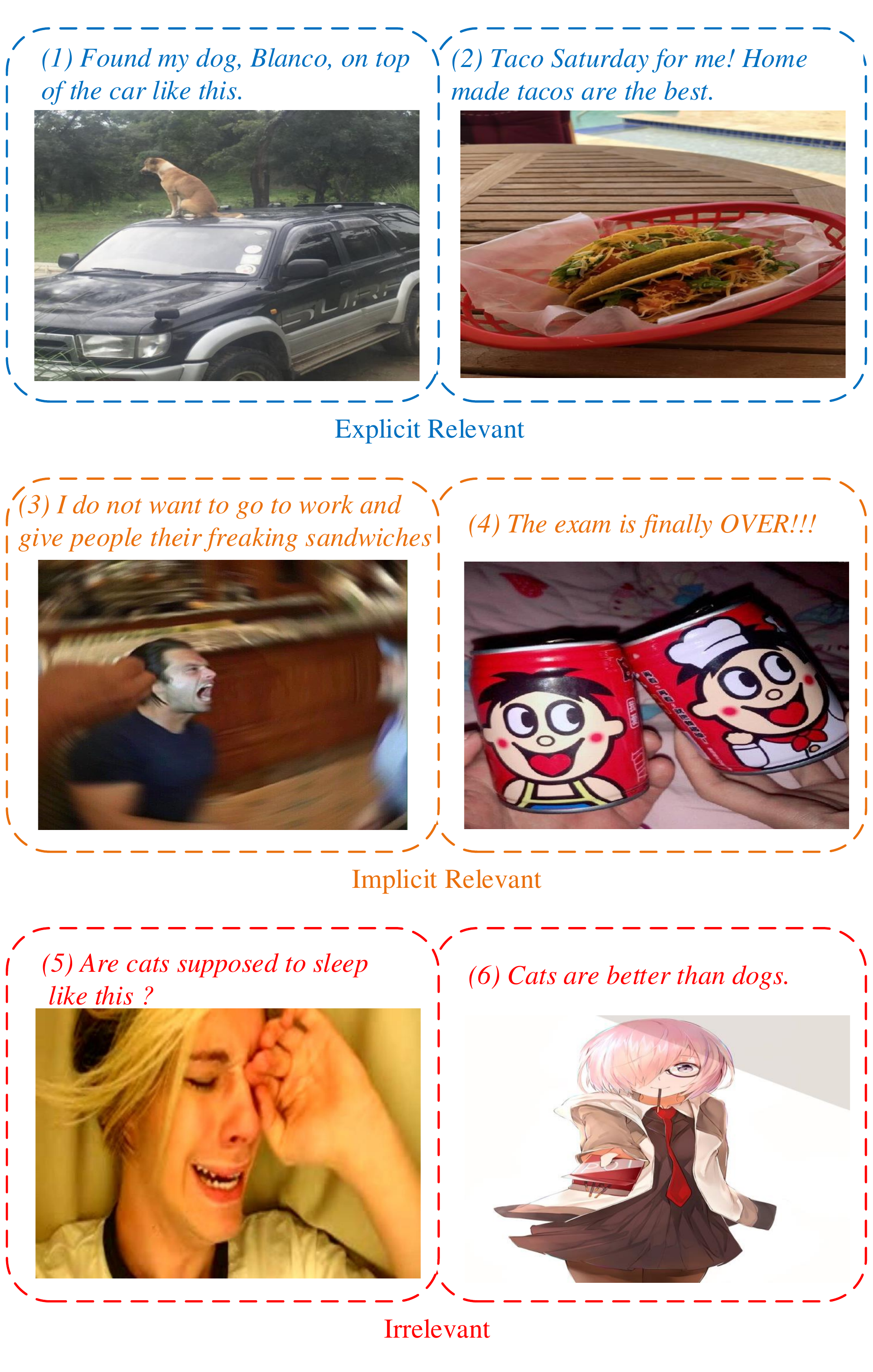}
	\caption{Examples of the three types of text-image correlation in this study. Explicit relevant tweets show detailed cross-modal alignments, such as \textit{dog} and \textit{car} in tweet (1), or \textit{tacos} in tweet (2); images and texts in implicit relevant tweets have none cross-modal alignment but are implicitly associated with each other based on implicit relevant, such as providing some relationship between shouting and protesting in tweet (3) or enhancing users' positive sentiment between image and text in tweet (4); images and texts in tweet (5) and (6)  are completely unrelated.}
	\label{fig:case}
\end{figure}

The category of cross-modal correlation has been studied in a few prior work. Early researchers focus on the definition of cross-modal correlation category and construct various classification systems according to multiple views, such as cross-modal contribution \cite{mccloud1998understanding}, cross-modal similarity \cite{marsh2003taxonomy}, logico–semantic and status relations \cite{martinec2005system}, visual distinction \cite{chen2013understanding}, cross-modal dependency \cite{wang2014bilateral}, systemic functional multimodal analyses \cite{wu2014multimodal}, etc. However, none of these studies propose any predictive models for cross-modal correlation categories. Recently, researchers pay more attention to the prediction of the cross-modal correlation categories and expand the existing classification system based on image specificity \cite{jas2015image}, emotion \cite{chen2015velda}, interrelation metrics \cite{henning2017,otto2020characterization}, parallel and non-parallel \cite{zhang2018equal}, contextual and semiotic relations \cite{kruk2019integrating}, visual content contribution \cite{vempala2019categorizing}, etc. They annotate data and train models to predict the cross-modal correlation category around the following tasks: multimodal regression \cite{jas2015image} or multimodal classification \cite{chen2015velda,vempala2019categorizing,henning2017,otto2020characterization,zhang2018equal,kruk2019integrating}.

Despite the landmark results of these predictive models, there are still two main drawbacks in the existing  studies about the definition and classification of cross-modal correlation. Firstly, existing definitions pay more attention to the fine-grained relevant types of cross-modal correlation, leading to a considerable part of implicitly relevant data is divided into irrelevant types according to the existing system. Take the tweet (3) and (4) in Fig.\ref{fig:case} as an example, these two image-text pairs will be classified to the category irrelevant according to the previous definition that \textit{image does not add to meaning} and \textit{text is not represented in image} \cite{vempala2019categorizing}. In fact, they have implicit associations based on some relationships (e.g. 'a shouting man' expresses protest against 'work' in Fig.\ref{fig:case}.3) or uses' sentiments (e.g. finished 'exam' is positively related to 'two cartoon smiling faces' In Fig.\ref{fig:case}.4). Secondly, previous predictive models combine the multimodal feature representations \cite{vempala2019categorizing}, or utilize cross-modal attention mechanism \cite{otto2020characterization} for image-text classification. But, none of them manifest the essence of cross-modal correlation according to their definition at the modeling stage. For example, the latest work is mainly centered on the role of the text to the tweet’s semantics and considers the image as the supplement to the text content, but it does not model this kind of inequality of modality importance in the proposed LSTM + InceptionNet model \cite{vempala2019categorizing}. 

In this paper, we focus on the most popular cross-modal data, image-text pair, and present a comprehensive analysis about the image-text correlation based on the implicit association and explicit alignment. According to our definition, it requires the predictive model to have the ability to reason about whether there are implicit associations or explicit alignments between the cross-modal situations. To support our argument, we create a new cross-modal correlation annotation protocol to construct a large scale image-text dataset and propose the Association and Alignment Network, namely AnANet. It consists of two parallel streams for visual and linguistic processing. The association net implicitly represents the global discrepancy and commonality between image and text via orthogonal decomposition. The alignment net explicitly captures the local relevance between image regions and text words via an interactive attention mechanism. The main contributions of our work are as follows:
\begin{itemize}
	\item We redefine a new classification system for image-text correlation based on implicit association and explicit alignment and propose a two-stream framework to model the essence of cross-modal correlation according to our definition.
	\item We construct the association and alignment network to represent the global implicit relevance and local explicit relevance between image and text, which allows for visiolinguistic distinguishable representations and enables sparse cross-modal interaction.
	\item We compare our model with the existing state-of-the-art methods, and the experimental results on our constructed new multimodal dataset demonstrate the effectiveness of our model in cross-modal correlation classification task.
\end{itemize}

\begin{table*}[htbp]
	\footnotesize
	\centering
\renewcommand\arraystretch{0.8}
\caption{Comparison of different cross-modal classification systems}
	\begin{tabular}{llm{32.625em}llm{8.125em}}
		\toprule
		Year  & Method & Category & Type  & Prediction & Domain \\
		\midrule
		1998  & McCloud‘s \cite{mccloud1998understanding} & 1.equal contributions to meanings of image and text\newline{}2.unequal contributions to meanings of image and text & Bool  & no    & Book \\
		\midrule
		2003  & Marsh's \cite{marsh2003taxonomy} & 1.image expressing little relation to the text\newline{}2.image expressing close relation to the text\newline{}3.image going beyond the text & Bool  & no    & Web Pages \\
		\midrule
		2005  & Martinec's \cite{martinec2005system} & 1.logico–semantic: expansion or projection\newline{}2.status relations: equal or unequal & Bool  & no    & Scientific Articles \\
		\midrule
		2013  & Chen's \cite{chen2013understanding} & 1.visually-relevant tweets\newline{}2.visually-irrelevant tweets & Bool  & no    & Social Media \\
		\midrule
		2014  & Wu's \cite{wu2014multimodal} & 1.elaboration\newline{}2.extension\newline{}3.enhancement\newline{}4.divergence & Bool  & no    & Book \\
		\midrule
		2014  & Wang's \cite{wang2014bilateral} & 1.independent image-text\newline{}2.image-text with similar meaning\newline{}3.text depending on image\newline{}4.image depending on text & Bool  & no    & Social Media \\
		\midrule
		2015  & Chen's \cite{chen2015velda} & 1.visually relevant\newline{}2.emotionally relevant\newline{}3.emotionally irrelevant & Bool  & yes   & Social Media \\
		\midrule
		2015  & Jas's \cite{jas2015image} & the specificity of given multiple descriptions with an image (i.e. 0.0-1.0) & Float & yes   & Caption \\
		\midrule
		2017  & Henning's \cite{henning2017} & 1.cross-modal mutual information\newline{}2.semantic correlation & Bool & yes   & Caption \\
		\midrule
		2018  & Zhang \cite{zhang2018equal} & 1.parallel\newline{}2.non-parallel & Bool  & yes   & Ad \\
		\midrule
		2019  & Kruk's \cite{kruk2019integrating} & 1.minimal\newline{}2.close\newline{}3.transcendent & Bool  & yes   & Social Media \\
		\midrule
		2019  & Kruk's \cite{kruk2019integrating} & 1.divergent\newline{}2.parallel\newline{}3.additive & Bool  & yes   & Social Media \\
		\midrule
		2019  & Vempala's \cite{vempala2019categorizing} & 1.image adds to the tweet meaning and text is represented in image\newline{}2.image adds to the tweet meaning and text is not represented in image\newline{}3.image does not add to meaning and text is represented in image\newline{}4.image does not add to meaning and text is not represented in image & Bool  & yes   & Social Media \\
		\midrule
		2020  & Otto's \cite{otto2020characterization} & 1.cross-modal mutual information\newline{}2.semantic correlation\newline{}3.status & Bool & yes   &Social Media,\newline{}Caption,\newline{}Wikipedia...\\
		\midrule
		2021  & Our   & 1.image and text are explicitly relevant\newline{}2.image and text are implicitly relevant\newline{}3.image and text are irrelevant & Bool  & yes   & Social Media \\
		\bottomrule
	\end{tabular}%
\label{tab:related}%
\end{table*}%

\section{Related Work}
This section will detailedly introduce the studies about the learning, definition, and classification of image-text correlation.

\subsection{Cross-modal Correlation Learning}
With the popularity of multimodal data in social media, more and more traditional tasks begin to process multimodal data based on cross-modal correlation learning. The image can provide rich visual semantic information for the text and assist in enhancing text content understanding, such as multimodel event classification \cite{ZEPPELZAUER201645}, multimodel topic labeling \cite{sorodoc2017multimodal}, multimodal named entity recognition \cite{zhang2018adaptive}, and multimodal hashtag recommendation \cite{ma2019co}. The theoretical insights of the connection between image and instructional text can be used to estimate automated discourse analysis \cite{alikhani2019cite}, guide image caption generation \cite{alikhani2020cross}, support multimodal summarization \cite{zhu2018msmo}, or learn cross-modal retrieval \cite{zhen2019deep}. The context-related information between the image and text can help analyze user's emotion \cite{zadeh2017tensor} or sentiment \cite{xu2018co}, and the context-conflicting information is also useful for multimodal sarcasm detection \cite{xu2020reasoning}. These works follow a prior assumption that the paired image and text are strictly related, but none of them explicitly define the specific categories of cross-modal correlation during the learning process. In other words, all these assumption that the text with an image represent similar concepts which are not true in social media (e.g. Twitter) \cite{vempala2019categorizing}.

\subsection{Cross-modal Correlation Definition}
Since the modeling of cross-modal correlation is the fundamental component of many multimodal applications, the definition of cross-modal correlation has gradually attracted the attention of researchers. The early definition of cross-modal correlation was established for specific data, such as six types of image-text relations for comic books which in terms of their equal/unequal contributions to meanings of image and text \cite{mccloud1998understanding}, three major categories for web pages according to the similarity between the illustration and text \cite{marsh2003taxonomy}, detailed classification system about image–text relations in scientific articles based on the logico–semantic and status relations \cite{martinec2005system}, a taxonomy of four image-text relationships in picture books based on systemic functional multimodal analyses \cite{wu2014multimodal}. In recent years, researchers focus on establishing classification systems for open source multimodal data, such as social media data \cite{chen2013understanding,wang2014bilateral}. Chen et al. \cite{chen2013understanding} uncover what people post about and the correlation between the tweet’s image and text, showing an important functional distinction between visually-relevant and visually-irrelevant tweets. Wang et al. \cite{wang2014bilateral} define the correspondences between image and text in microblog on the basis of their dependency conditions (i.e. dependent or independent), to assist the discovery of multimodal social topics. However, the above research mainly focuses on the construction of classification systems for image-text correlation, none of them propose such predictive models for image-text correlation categories.

\subsection{Cross-modal Correlation Classification}
According to the previous definition of image-text correlation, researchers recently pay attention to the prediction of the correlation categories. The early work finds that both visual elements and emotional elements play important roles in image-text correlation \cite{chen2015velda}, and develop a visual emotional topic model to capture the image-text correlation from visual or emotional perspectives. To measure the similarity between an image with multiple description texts, Jas and Parikh \cite{jas2015image} propose the concept of image specificity and turn the cross-modal correlation classification problem into a specificity regression problem. Zhang et al. \cite{zhang2018equal} focus on the parallel and non-parallel relationships between image and text slogan and design 9 features (e.g. topics, slogan specificity and concreteness, etc.) to train an ensemble of SVM classifiers. To explore contextual relations (i.e. minimal, close, or transcendent) and semiotic relations (i.e. divergent, additive, or parallel) behind image-text pair, two taxonomies are also proposed to capture literal and semiotic meanings of image-text pair \cite{kruk2019integrating}. According to the role of the image to the semantics of the tweet, Vempala and Preoţiuc-Pietro \cite{vempala2019categorizing} aim to identify if the image’s content contributes with additional information to the meaning of the tweet beyond the text via jointing image-text neural networks. Although the above researches have proposed a variety of image-text correlation classification methods, they still treat this task as the standard image-text classification task via simply combining the multimodal feature encoders for image-text classification at the prediction stage, such as SVM \cite{zhang2018equal}, InceptionNet+LSTM \cite{vempala2019categorizing}, ResNet+GRU \cite{kruk2019integrating}. In fact, the joint encoding between image and text is very helpful for cross-modal modeling \cite{henning2017, otto2020characterization}. To describe different interrelations of textual and visual information, Heening and Ewerth \cite{henning2017} propose two metrics, namely cross-modal mutual information(CMI) and semantic correlation(SC). They utilize an Multimodal Autoencoder (InceptionNet+LSTM) to gather a compact embedding, where the first input of the text LSTM layer is the image embedding in order to emulate a natural article processing (reading the text under consideration of the enclosed image). Furthermore, Otto et al. \cite{otto2020characterization} construct a new taxonomy of eight semantic image-text relationships via leveraging another metric describing status relation of image-text pair into  CMI and SC \cite{henning2017} and propose a image guided textual attention layer to ensure that the neural network reads the textual information under the consideration of the visual features, which forces it to interpret the features in unison. Only a small amount of work has considered the interactive relationship between graphics and text in the process of modeling graphics and text embedding

The Table \ref{tab:related} shows a detailed comparison of existing cross-modal classification systems. There are two main types of image-text correlation, relevant and irrelevant, and the existing work pays more attention to fine-grained relevant correlation classification. However, in practical applications, a considerable part of the data is divided into the irrelevant types according to the existing system which is actually implicitly relevant. And the worst is that, for those predictive methods, little is known about how textual content is related to the images with which they appear. That means none of them manifest the essence of cross-modal correlation according to their definition at the modeling stage. Thus, in this paper, we first define an annotation scheme that focuses on the explicit or implicit correlation between image and text and then propose a multimodal classifier to model cross-modal correlation according to our definition.

\section{Definition and Data Collection}
\subsection{New Definition of Cross-modal Correlation}
Different from the various definitions of cross-modal correlation in previous work which pay more attention to the fine-grained relevant types of cross-modal correlation, in this paper, we focus on finding those implicitly relevant data which are often divided into irrelevant types according to the existing classification systems.

We focus on the most popular cross-modal data, image-text pair, and define the new classification system of image-text correlation based on implicit association and explicit alignment, including explicit relevant, implicit relevant, and irrelevant. To research the cross-modal correlation and support our argument, we create a new cross-modal correlation annotation protocol according to our definition, which uses the following guidelines:

\textbf{Explicit relevant.} The image and text have the same meaning with explicitly alignments:
\begin{itemize}
	\item Some or all of the visual regions are depicted in the text (e.g. Fig.\ref{fig:case}.1).
	\item Some or all of the textual words are shown in the image (e.g. Fig.\ref{fig:case}.2).
\end{itemize}

\textbf{Implicit relevant.} The image and text have the different meanings, but:
\begin{itemize}
	\item One modality makes a reference to something or depicts something that adds information to another modality (e.g. Fig.\ref{fig:case}.3).
	\item One modality enhances the sentiment or expresses a feeling about the content of another modality (e.g. Fig.\ref{fig:case}.4).
\end{itemize}

\textbf{Irrelevant.} The information of each modality is independent and does not add any useful content to represent the whole meaning of the image-text pair (e.g. Fig.\ref{fig:case}.5 and Fig.\ref{fig:case}.6).

\begin{table}[tbp]
	\centering
	\caption{Statistics of our constructed dataset}
	\renewcommand\arraystretch{1}
	\renewcommand\tabcolsep{12pt}
	\begin{tabular}{ccccc}
		\toprule[1.3pt]
		& Train & Dev   & Test  & All \\
		\midrule
		Explicit Relevant & 5779  & 542   & 558   & 6879 \\
		Implicit Relevant & 1070  & 110   & 106   & 1286 \\
		Irrelevant & 3445  & 323   & 319   & 4087 \\
		\midrule
		All   & 10294 & 975   & 983   & 12252 \\
		\bottomrule[1.3pt]
	\end{tabular}%
	\label{tab:data}%
\end{table}%

\subsection{Data Collection and Annotation}
In this paper, we focus the text-image relationship in social media. The latest research studies how the meaning of the entire tweet is composed through the relationship between its textual content and its image and conduct a Text-Image Relationship dataset \cite{vempala2019categorizing}, including 4,471 multimodal tweets. We expand this dataset and also use Twitter as our data source to collect more image-text pairs, getting a larger scale Cross-modal Correlation Dataset, namely CCD (will be released soon). 

We hired 3 annotators to judge the type of each image-text pair independently, and also use the majority vote strategy \cite{vempala2019categorizing} to aggregate those judgments as the final label. We randomly divide this dataset into the training set, development set and test set, and the detailed statistics are given in Table \ref{tab:data} .

\begin{figure*}[ht]
	\centering
	\includegraphics[width=1\linewidth]{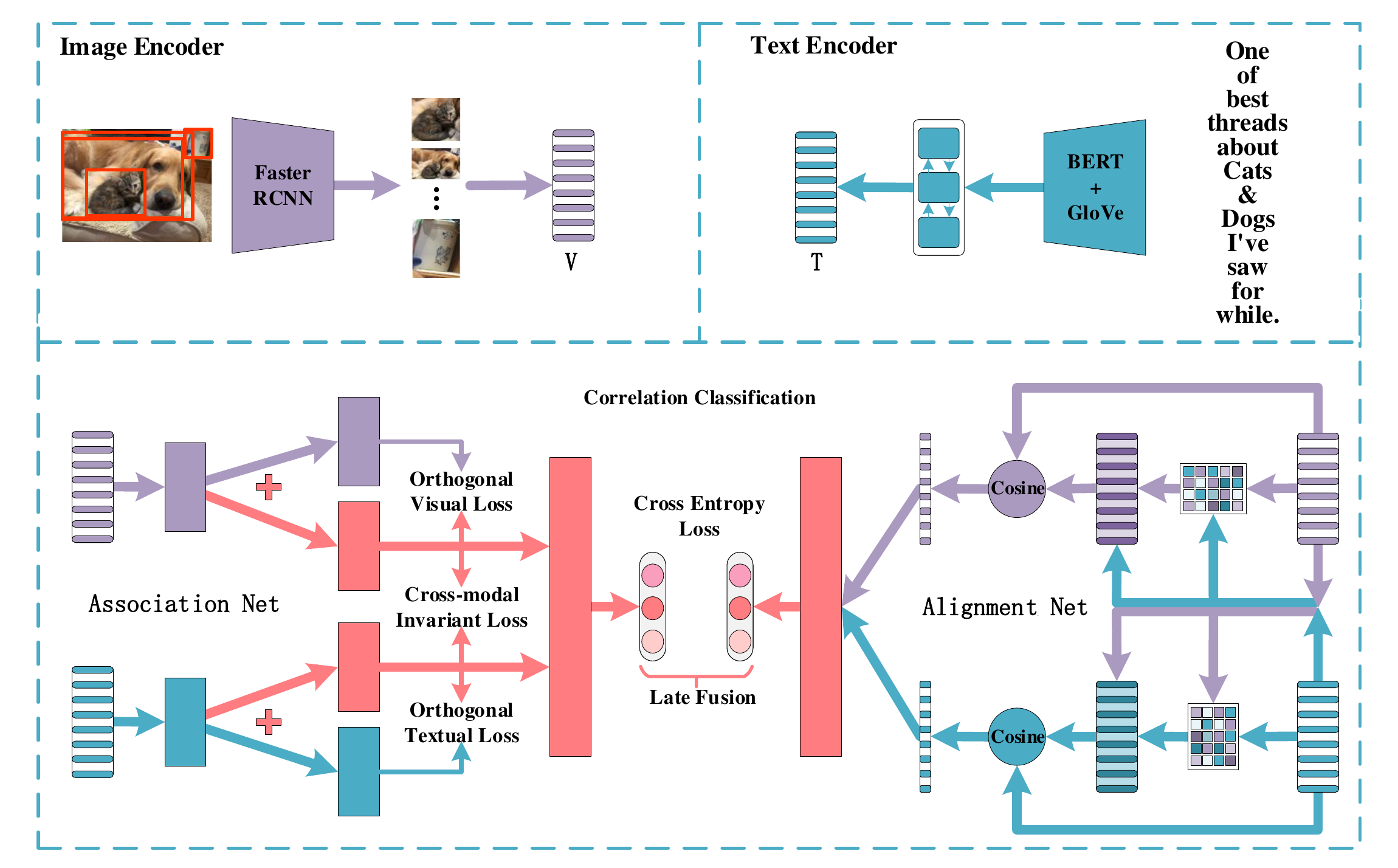}
	\caption{Overall architecture of our proposed AnANet for cross-modal correlation classification. 
	}
	\label{fig:model}
\end{figure*}

\section{Proposed Model}
Fig.\ref{fig:model} illustrates the overall architecture of our proposed model for cross-modal correlation classification. We first learn the single-modality representations via two independent encoders, and then design the AnANet to model the cross-modal correlation through two parallel streams: an association net to globally weaken the discrepancy and strengthen the commonality; a alignment net to find relevant anchors to capture cross-modal local relevance.

\subsection{Encoders}
Given an image-text pair $(Image, Text)$, we first use two independent encoders to map them into single-modality feature representations: a set of image features $V=\{v_1,v_2,\cdots,v_K\} \in \mathbb{R}^{K \times d}$ and a set of word features $T=\{t_1,t_2,\cdots,t_N\} \in \mathbb{R}^{N \times d}$, where each image feature $v_i$ encodes a region in the $Image$, each word feature $t_j$ represents the word $w_j$ in the $Text$, $K$ and $N$ are the lengths of image region sequence and text word sentence, $d$ is the dimensions of the encoded image and text features.

\textbf{Image Encoder} uses the famous image object detection framework, Faster RCNN \cite{2015Faster}, to select $K$ image regions and encode them as sequential features $V=\{v_1,v_2,\cdots,v_K\}$. Unlike prior work \cite{vempala2019categorizing} directly utilizing InceptionNet to extract the output of the last fully-connected layer as the global image feature, this structure allows for more detailed local objects and enables more visual semantic information.
\begin{gather}
	V=\sigma(w_v*Faster RCNN(Image)+b_v) \in \mathbb{R}^{K \times d}
\end{gather}
where $\sigma(\cdot)$ is the $ReLU$ activation function, $w_v,b_v$ are the parameters of the fully-connected layer.

\textbf{Text Encoder} transforms each word token $w_j$ in the text into continuous representation via both word embedding $w_j^G$ and contextualized embedding $w_j^B$. The former is embedded by GloVe \cite{2014GloVe} pretrained on large social media corpus, and the latter is embedded by BERT \cite{devlin-etal-2019-bert} which leverages prior linguistic knowledge and has been shown success in capturing syntactic, semantic, and world knowledge. We concatenate these two embeddings and utilize the Bi-directional GRU layer to model the contextual information and generate the final word features $T$.
\begin{gather}
	w_j^G=Embed_{GloVe}(w_j) \in \mathbb{R}^{d_G} \\
	w_j^B=Embed_{Bert}(w_j)\in \mathbb{R}^{d_B} \\
	T=BiGRU([w_j^G \oplus w_j^B]) \in \mathbb{R}^{N \times d}
\end{gather}
where $\oplus$ denotes the operation of vector concatenation, $d_G$ and $d_B$ are the dimensions of word embedding and contextualized embedding respectively.

\subsection{Association Net} 
To focus on the implicit association between image and text, we design the association net to implicitly decompose the image and text into cross-modal discrepancy and commonality, and then reinforce this kind of commonality as the global relevant representation.

\textbf{Cross-modal Decomposition.} It consists of a common layer, visual decomposition layer, and textual decomposition layer. We feed the overall visual and textual features $f^*$ into these three layers and break down them into shared invariant features $f^*_{inv}$ and unique variant features $f^*_{var}$. 
\begin{gather}
	f^*_{inv} = Wf^* \in \mathbb{R}^{d_{inv}}\\
	f^*_{var}  = P^{*}f^* \in \mathbb{R}^{d_{var}}
\end{gather} 
where $*$ indicates the input modality $image$ or $text$, , $f^{image}$ is the average of image feature sequence $V$, $f^{text}$ is the average of text feature sequence $T$. To ensure the conflict of different subspaces, we further enforce an additional orthogonal constraint on the unique modality projection matrices $P^{*} \in \mathbb{R}^{d_{var}\times d}$ and the shared projection matrix $W \in \mathbb{R}^{d_{inv}\times d}$.
\begin{gather}\label{cons}
	W^TP^*=0 \ (*\in\{image, text\})
\end{gather}

\textbf{Commonality Strengthen.} We regard the shared invariant features $f^*_{inv}$ as the representation of cross-modal commonality, and the unique variant features $f^*_{var}$ as the representation of cross-modal discrepancy. In cross-modal correlation classification task, we expect our model to focus more on the relevance between image and text. Thus, we reinforce their cross-modal commonality and weaken their cross-modal discrepancy. Specifically, we abandon the unique variant features and directly concatenate the above shared invariant features as the global relevant representation $r_{g}$.
\begin{gather}
	r_{g} = [f^{image}_{inv}\oplus f^{text}_{inv}]\in \mathbb {R}^{2d_{inv}} 
\end{gather} 

\subsection{Alignment Net}
To further represent the explicit alignments between different modality information, we design the alignment net to absorb image features $V$ and text features $T$, and infer the local relevant anchors by aligning image region and text word. 

\textbf{Interactive Attention.} Our interactive attention mechanism firstly calculates the similarity matrix to measure how image regions and text words relate, which is defined with two complementary formulations below.
\begin{gather}
	A_{t2v} = Att(V,T) \in \mathbb{R}^{K \times N}\\
	A_{v2t} = Att(T,V) \in \mathbb{R}^{N \times K}
\end{gather}
where the attention function $Att(\cdot)$ specified by the input matrix $H\in \mathbb{R}^{L \times d}$ and $Q\in \mathbb{R}^{M \times d}$ is defined as:
\begin{gather}
	Att(H,Q)=\left[\frac{\exp \left(s_{i j}\right)}{\sum_{i} \exp \left(s_{i j}\right)}\right]_{i, j} \in \mathbb{R}^{L \times M}\\
	s_{ij} = H_{i:}Q_{:j}^{T}
\end{gather}

To attend on words with respect to each image region, we calculate the weighted combination of text word features $T$ as the text guided visual attention feature $\hat{V}$.
\begin{gather}
	\hat{V}=A_{t2v}*T \in \mathbb{R}^{K \times d}
\end{gather}

Similarly, we also calculate the weighted average of image region features $V$ according to the attention matrix $A_{v2t}$ to get the image guided textual attention feature $\hat{T}$.
\begin{gather}
	\hat{T}=A_{v2t}*V \in \mathbb{R}^{N \times d}
\end{gather}

\textbf{Pair-wise Comparison.} After the interactive attention layer, the attended multimodal features $\hat{V}$ and $\hat{T}$ have aggregated the one modality information with respect to another modality information. Though the strategy of directly combining multimodal features is very common in existing multimodal work \cite{zhang2018adaptive,vempala2019categorizing}, it can not measure the degree of local association between image regions and text words in this task. Hence, we design the pair-wise comparison strategy to quantify the cross-modal local relevance. The pair-wise comparison function $f(\cdot)$ uses cosine function to compare the original modality feature $V$ and $T$ with the cross-modal emphasized attention feature $\hat{V}$ and $\hat{T}$ term by term.
\begin{align}
	\overline{v} = f(V,\hat{V}) =[cosine(V_{i:},\hat{V}_{i:})]_{i=1}^{K} \in \mathbb{R}^{K}\\
	\overline{t} = f(T,\hat{T}) =[cosine(T_{i:},\hat{T}_{i:})]_{i=1}^{N} \in \mathbb{R}^{N}
\end{align}

This structure enables our model to measure the degree of local relevance between image regions and text words according to the similarity items pointed by the two indicator sequences $\overline{v}$ and $\overline{t}$. Then, we concatenate these sparse cross-modal interaction features as the local relevant representation $r_{l}$ for cross-modal correlation classification.
\begin{gather}
	r_{l} = [\overline{t}\oplus\overline{v}] \in \mathbb{R}^{K+N}
\end{gather} 

\subsection{Classification}
Due to the heterogeneity of global and local cross-modal relevant representations $r_{g}$ and $r_{l}$, we utilize the late fusion strategy for final classification. It first uses two fully-connected layers to predict the correlation type of image-text respectively, and then aggregates these two results as the final prediction $\hat{y}$.
\begin{gather}\label{y}
	\hat{y} = softmax\left(\lambda(w_{g}r_g+b_g)+\eta(w_{l}r_l+b_l)\right)
\end{gather} 
where $\lambda$ and $\eta$ are hyperparameters to control the contributions of association net and alignment net for cross-modal correlation classification.

\begin{table*}[tbp]
	\centering
	\renewcommand\arraystretch{1}
	\renewcommand\tabcolsep{20pt}
	\caption{Comparative results with multimodal baselines}
	\begin{tabular}{cllcc}
		\toprule[1.3pt]
		\textbf{Group}& \textbf{Method} & \textbf{Encoder} & \textbf{Acc } & \textbf{F1} \\
		\midrule
		\multirow{2}[2]{*}{1} & Random & -     & 35.45  & 38.30  \\
		& Majority \cite{vempala2019categorizing} & -     & 56.19  & 40.43  \\
		\midrule
		\multirow{3}[2]{*}{2} & TFN \cite{zadeh2017tensor} & FN, LSTM(GloVe) & 79.76  & 79.46  \\
		& CoMemory \cite{xu2018co} & InceptionV3, Memory Network(GloVe) & 77.82  & 77.42  \\
		& SCAN \cite{lee2018stacked} & Faster RCNN, GRU(GloVe) & 80.87  & 79.49  \\
		\midrule
		\multirow{4}[2]{*}{3} & InceptionNet+LSTM \cite{vempala2019categorizing} & InceptionNet, LSTM(GloVe) & 76.93  & 75.45  \\
		& DCNN(GloVe) \cite{kruk2019integrating} & ResNet18, GRU(GloVe) & 79.87  & 78.92  \\
		& DCNN(GloVe+BERT)  \cite{kruk2019integrating}  & ResNet18, GRU(GloVe+BERT) & 82.70  & 82.34  \\
		& Deep Classifier \cite{otto2020characterization} & InceptionV2, GRU(GloVe) & 80.47  & 79.14  \\
		\midrule
		\multicolumn{1}{c}{\multirow{2}[2]{*}{Ours}} & AnANet(GloVe) & Faster RCNN, GRU(GloVe) & 81.58  & 81.09  \\
		& AnANet(GloVe+BERT) & Faster RCNN, GRU(GloVe+BERT) & \textbf{84.84} & \textbf{83.84} \\
		\bottomrule[1.3pt]
	\end{tabular}%
	\label{tab:main}%
\end{table*}%

\subsection{Optimization}
We optimize our model with three losses: classification loss, cross-modal invariant loss, and modality orthogonal loss.

\textbf{Classification Loss.} To optimize the final classification layer, we use the cross entropy loss function which is the most commonly used in classification tasks as the cross-modal correlation classification loss $\mathcal{L}_{c}$.
\begin{gather}\label{lc}
	\mathcal{L}_{c}=-\sum_{i} y _{i} \log{\hat{y}_i}
\end{gather} 
where $y_i$ is the ground truth of $i$-th image-text pair (i.e. 0 for irrelevant, 1 for implicit relevant, and 2 for explicit relevant), and $\hat{y}_i$ is the predicted label of our model.

\textbf{Cross-modal Invariant Loss.} In association net, we share the same matrix $W$ to ensure projecting image and text into the same subspace and regard the shared invariant features $f^*_{inv}$ as the representation of cross-modal commonality. To ensure the similarity of decomposed shared invariant features, we propose the cross-modal invariant loss $\mathcal{L}_{i}$, denoted as:
\begin{gather}\label{li}
	\mathcal{L}_{i}=\left \| f^{image}_{inv} -f^{text}_{inv}\right \|_2
\end{gather} 
where $\left \| \cdot \right \|_2$ denotes the 2 norm.

\textbf{Modality Orthogonal Loss.} In association net, we also use two unique modality projection matrices $P^{image}$ and $P^{text}$ to decompose the image and text into unique variant subspace under the additional orthogonal constraint (Eq.(\ref{cons})). Here, we convert this constraint into the following modality orthogonal loss $\mathcal{L}_{o}$.
\begin{gather}\label{lo}
	\begin{split}
		\mathcal{L}_{o} = & \sum_{*\in\{image, text\}} \left \| W^TP^* \right \|_F \\
		{\bf s.t}\quad & W^TP^{*} =0 \ (*\in\{image, text\})\\
	\end{split}
\end{gather}
where $\left \| \cdot \right \|_F$ denotes the Frobenius norm.

According to Eq.(\ref{lc}-\ref{lo}), we finally minimize the combined loss function to optimize our whole network:
\begin{gather}\label{l}
	\mathcal{L}=\mathcal{L}_{c}+\alpha \mathcal{L}_{i}+\beta\mathcal{L}_{o}
\end{gather} 
where $\alpha$ and $\beta$ are hyperparameters to control the distributions of the cross-modal invariant loss and modality orthogonal loss.

\section{Experiment}
\subsection{Implementation Details} 
We evaluate the proposed AnANet on our constructed CCD dataset for cross-model correlation classification. Each sample in this dataset is an image-text pair. For each image, we first resize it into 224*224 and utilize pretrained Faster RCNN\footnote{https://github.com/jwyang/faster-rcnn.pytorch} to select 36 image regions and encode them into 1024 image region features. For each text, we embed each word into 200-dimensional word embedding by GloVe\footnote{https://github.com/stanfordnlp/GloVe} pretrained on the Twitter data and 768-dimensional contextual embedding by BERT\footnote{https://github.com/google-research/bert}. We set the maximum length of the text sequence to 100 and encode the text into 1024-dimensional hidden space via BiGRU. We map multimodal features into three projection subspaces and set the dimensions of them to 200. We optimize our model by Adam update rule with 64 mini-batch and 0.001 learning rate.

\subsection{Baseline Methods}
We compare our model with the following multimodal baseline methods, including the lastest methods in cross-modal correlation classification task (i.e. InceptionNet+LSTM, DCNN, Deep Classifier), several representative methods in multimodal learning tasks which are modified for multimodal classification (i.e. TFN, CoMemory, SCAN), and two simple baseline methods (i.e. Random and Majority). 

\begin{itemize}
	\item \textbf{Random} predicts the cross-modal correlation category of the image-text pair randomly.
	\item \textbf{Majority} always predicts the most frequent class in experimental dataset (i.e. explicit relevant category).
	\item \textbf{InceptionNet+LSTM} \cite{vempala2019categorizing} concatenates multimodal features generated by InceptionNet and LSTM for multimodal classification. 
	\item \textbf{Deep Classifier} \cite{otto2020characterization} force the the GRU model to read the textual information under the consideration of the visual
	features via a image guided attention mechanism. 
	\item \textbf{DCNN} \cite{kruk2019integrating} takes the ResNet-18 and bidirectional GRU for cross-modal correlation classification, which utilizes both word2vec type and pre-trained character-based contextual embeddings (ELMo) for textual embedding. For fair comparison, we replace its ELMo embedding with the stronger language model BERT used in our model.
	\item \textbf{TFN} \cite{zadeh2017tensor} uses a joint tensor fusion network to model the intra-modality and inter-modality dynamics of the multiple information. 
\item \textbf{CoMemory} \cite{xu2018co} models the interactions between image and text iteratively via co-memory mechanism. 
\item \textbf{SCAN} \cite{lee2018stacked} discovers the full latent alignments using both image regions and words to infer image-text similarity via stacked cross attention. 
\end{itemize}

\begin{table*}[ht]
	\centering
	\renewcommand\arraystretch{1}
	\caption{Ablation Study: Different Components of AnANet}
	\begin{tabular}{rlllcccc}
		\toprule[1.3pt]
		\multicolumn{2}{l}{\textbf{Method}} & \textbf{Module} & \textbf{Encoder} & \textbf{Acc } & \textbf{F1} & \textbf{$\triangle$Acc } & \textbf{$\triangle$F1} \\
\midrule
		1     & AnANet & Association Net, Alignment Net & Faster RCNN, GRU(GloVe+BERT) & \textbf{84.84 } & \textbf{83.84 } & -     & - \\
\midrule
2     & ~~  -BERT & Association Net, Alignment Net & Faster RCNN, GRU(GloVe) & 81.58  & 81.09  & -3.26  & -2.75  \\
3     & ~~  -Association Net & Alignment Net & Faster RCNN, GRU(GloVe+BERT) & 83.21  & 82.32  & -1.63  & -1.52  \\
4     & ~~  -BERT & Alignment Net & Faster RCNN, GRU(GloVe) & 80.87  & 79.49  & -3.97  & -4.35  \\
5     & ~~  -Alignment Net & Association Net & Faster RCNN, GRU(GloVe+BERT) & 83.52  & 83.20  & -1.32  & -0.64  \\
6     & ~~  -BERT & Association Net & Faster RCNN, GRU(GloVe) & 81.18  & 80.09  & -3.66  & -3.75  \\
7     & ~~  -Association Net, -Alignment Net & AvgPooling, Concatnation & Faster RCNN, GRU(GloVe+BERT) & 83.01  & 82.28  & -1.83  & -1.56  \\
8     & ~~  -BERT & AvgPooling, Concatnation & Faster RCNN, GRU(GloVe) & 81.28  & 80.18  & -3.56  & -3.66  \\
		\bottomrule[1.3pt]
	\end{tabular}%
	\label{tab:embedding}%
\end{table*}%

\subsection{Main Result}
We use the accuracy and weighted F1-score as our experimental metrics to evaluate the performance of different methods. Table \ref{tab:main} shows the comparative results on our constructed dataset and our AnANet achieves the best performance among all compared baseline methods.

By comparing baseline methods in cross-modal correlation classification task (Group 3 of Table \ref{tab:main}), 
our AnANet(GloVe) performs better than those two combination based methods, DCNN(GloVe) and InceptionNet+LSTM, which extract multimodal features via combining different encoders. It indicates that our proposed association net and alignment net  can effectively model the cross correlation and learn high-quality multimodal representations. Deep Classifier performs better than DCNN(GloVe) and InceptionNet+LSTM but is also weaker than our AnANet(GloVe), which shows that the image guided attention attention mechanism used in Deep Classifier is not enough to capture the bidirectional correlation between image and text, which is modeled by our interactive attention of alignment net. It is important that when we introduce the strong language model BERT into DCNN and our AnANet models, they achieve great performance improvements and our AnANet(GloVe+BERT) performs better than all cross-modal correlation classification baselines. It demonstrates that BERT model which has learned common language representations via pre-training on large-scale corpus is very helpful for downstream tasks, including cross-modal correlation classification task.

We also compare our model with those representative multimodal classification methods in Group 2 of Table \ref{tab:main}. The SCAN method outperforms all baseline methods, indicating the selection of image encoder is crucial for this task. The SCAN method and our AnANet(GloVe) both use the Faster RCNN as image encoder, which is able to discover the local clues between the image region and the text word than global encoder used in other baseline methods (e.g. InceptionNet, ResNet, FN, etc.). Further, compared with SCAN which also utilizes the interactive attention architecture to capture cross-modal correlation and infers the image-text similarity via modeling the full latent alignments of image regions and words (similar to our alignment net), our AnANet still performs better. It is because that we also introduce an extra association net to focus more on the commonality between image and text and weaken their cross-modal discrepancy, which further indicates the effectiveness of our two-stream framework.

\begin{figure*}
	\centering
	\includegraphics[width=1\linewidth]{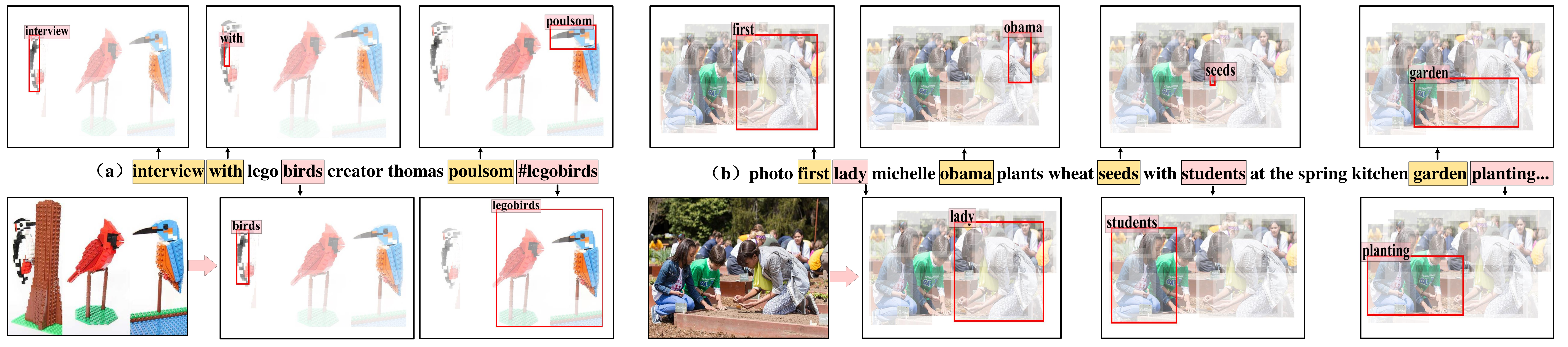}
	\caption{Visualization of the attended image regions with respect to related word in different sub-figures. We use different regional brightness to represent the attentional weight of both region and word estimated by our interactive attention layer, and outline the maximum attentional region of each word in red.}
	\label{fig:att}
\end{figure*}

\subsection{Ablation Study}
In this section, we carry out extensive ablation studies on different variants of AnANet to evaluate the performance of its components.
\subsubsection{Different Components of AnANet} 
The detailed ablation results about different components of AnANet are shown in Table \ref{tab:embedding}. The most obvious declines come from the direct removal of BERT model (see rows 2, 4, 6, 8). Each of these variants which only use the GloVe word vector in the text has a significant performance degradation, indicating that the pre-trained language model can provide better word representations for the downstream task and bring about 3.26\% accuracy gain and 2.75\% F1 gain on this task. Comparing these two core modules (i.e. Association Net and Alignment Net) used in our model, we find that removing Association Net has greater performance drop than removing Alignment Net (see rows 3, 5), indicating global implicit association captured by Association Net plays a more important role than local explicit alignment extracted by Alignment Net for this task. We also remove all modules proposed in this paper and directly concatenate the multimodal representations for classification (see rows 7, 8), similar to those combination based  baselines (e.g. DCNN, InceptionNet+LSTM, TFN), these variants of our AnaNet are still dominant, further demonstrating the superiority of Faster RCNN as the image encoder for depicting the local details of the image in this task.

\begin{table}[tbp]
	\centering
	\caption{Comparative results with different modules}
		\renewcommand\arraystretch{1}
			\renewcommand\tabcolsep{9pt}
	\begin{tabular}{lccccc}
		\toprule
		\textbf{	{Method}} & Acc &F1 & \textbf{{$F_{exr}$}} &\textbf{{$F_{imr}$}} & \textbf{{$F_{irr}$} }\\
		\midrule
		AnANet & 84.84  & 83.84  & 90.22  & 42.86  & 87.97  \\
Association Net & 83.52  & 83.20  & 90.02  & 43.65  & 86.82  \\
Alignment Net & 83.21  & 82.32  & 89.68  & 37.71  & 86.59  \\
		\bottomrule[1.3pt]
	\end{tabular}%
	\label{tab:f1}%
\end{table}%

\subsubsection{Module Bias for Different Types of Data}
In this paper, we introduce a new classification system of image-text correlation, and design a two-stream framework according to our definition, expecting it to be good at identifying such kinds of data. In this section, we further explore the preferences of the two core modules of our model for different categories of data. Detailed comparison results are shown in Table \ref{tab:f1}. {$F_{exr}$}, {$F_{imr}$} and {$F_{irr}$} denote the F1 scores of explicit relevant, implicit relevant, irrelevant classes respectively. It can be seen that the two modules have the same ability to identify explicit relevant samples and irrelevant samples. But, for the implicit correlation samples, Association Net shows strong dominance, reaching 43.65\% in {$F_{imr}$}, far surpassing Alignment Net's 37.71\%. Being good at identifying this kind of implicit relevant sample is an important factor of our model's superiority over other kinds of baseline methods.

\subsubsection{Different Feature Fusions of Association Net}
We also conduct the ablation studies on different features used in Association Net to evaluate the effectiveness of our commonality strengthen operation. The detailed results of different feature combination strategies are shown in Table \ref{tab:x1}. In various permutations and combinations of all features, only using the shared invariant features $f^{image}_{inv}, f^{text}_{inv}$ achieves the best performance (see row 1). 
It is obvious that using our decomposed invariant or variant features for multimodal feature fusion is more effective than directly using the original image and text features (see rows 1, 2, 3). It also works well when we combine these decomposed features with the original features (see rows 3, 4, 5). The results indicate that our Association Net has the ability to learn more distinguishable and high-quality multimodal features. Interestingly, when we combine invariant features and variant features for feature fusion (see row 6), it seems that they are mutually exclusive and even better to directly use original image and text features. Moreover, it is not that the more features used, the stronger the representation ability of the model will be, which not only confuses the model but also increases its complexity (see rows 4, 5, 6, 7).

\begin{table}[tbp]
	\centering
	\footnotesize
	\renewcommand\arraystretch{1.1}
	\renewcommand\tabcolsep{1.4pt}
	\caption{Ablation Study: Different Feature Fusions of Association Net}
	\begin{tabular}{llcccc}
		\toprule[1.3pt]
		\multicolumn{2}{l}{\textbf{Variant}} & \textbf{Acc } & \textbf{F1} & \textbf{$\triangle$Acc} & \textbf{$\triangle$F1} \\
		\midrule
		1     & $f^{image}_{inv},f^{text}_{inv}$ (AnANet Used) & 84.84  & 83.84  & -     & - \\
\midrule
2     & $f^{image}_{var}, f^{text}_{var}$ & 84.02  & 83.49  & -0.82  & -0.35  \\
3     & $ f^{image}, f^{text}$ & 83.92  & 82.84  & -0.92  & -1.00  \\
4     & $f^{image}_{var}, f^{text}_{var}, f^{image}, f^{text}$ & 84.02  & 83.38  & -0.82  & -0.46  \\
5     & $f^{image}_{inv}, f^{text}_{inv}, f^{image}, f^{text}$ & 84.13  & 83.36  & -0.71  & -0.48  \\
6     & $f^{image}_{var}, f^{image}_{inv}, f^{text}_{inv}, f^{text}_{var}$ & 83.32  & 83.05  & -1.52  & -0.79  \\
7     & $f^{image}_{var}, f^{image}_{inv}, f^{text}_{inv}, f^{text}_{var}, f^{image}, f^{text}$ & 83.42  & 82.91  & -1.42  & -0.93  \\
		\bottomrule[1.3pt]
	\end{tabular}%
	\label{tab:x1}%
\end{table}%

\section{Visualization and Analysis}
\subsection{Visualizing Attention}

Our proposed Alignment Net utilizes interactive attention mechanism to explicitly calculate the local relevance between image regions and text words. By visualizing the attention component learned by our model on two relevant multimodal tweets in Fig.\ref{fig:att}, we are able to provide the interpretability of our Alignment Net. We can observe that the words "birds", "legobirds", "lady", "students" and "planning" receive strong attention on the relatively precise locations, but other words, like "interview", "with", "poulsom", "first", "obama", "seed", "garden" etc., are less focused in sub-figures. The results show that our proposed attention mechanism works well for cross-modal correlation classification via providing the explainable reasoning cross-modal alignments between image regions and text words.

\subsection{Hyperparameter Analysis}
\begin{figure}[tbp]
	\centering
	\includegraphics[width=1\linewidth,trim=38 30 60 60,clip]{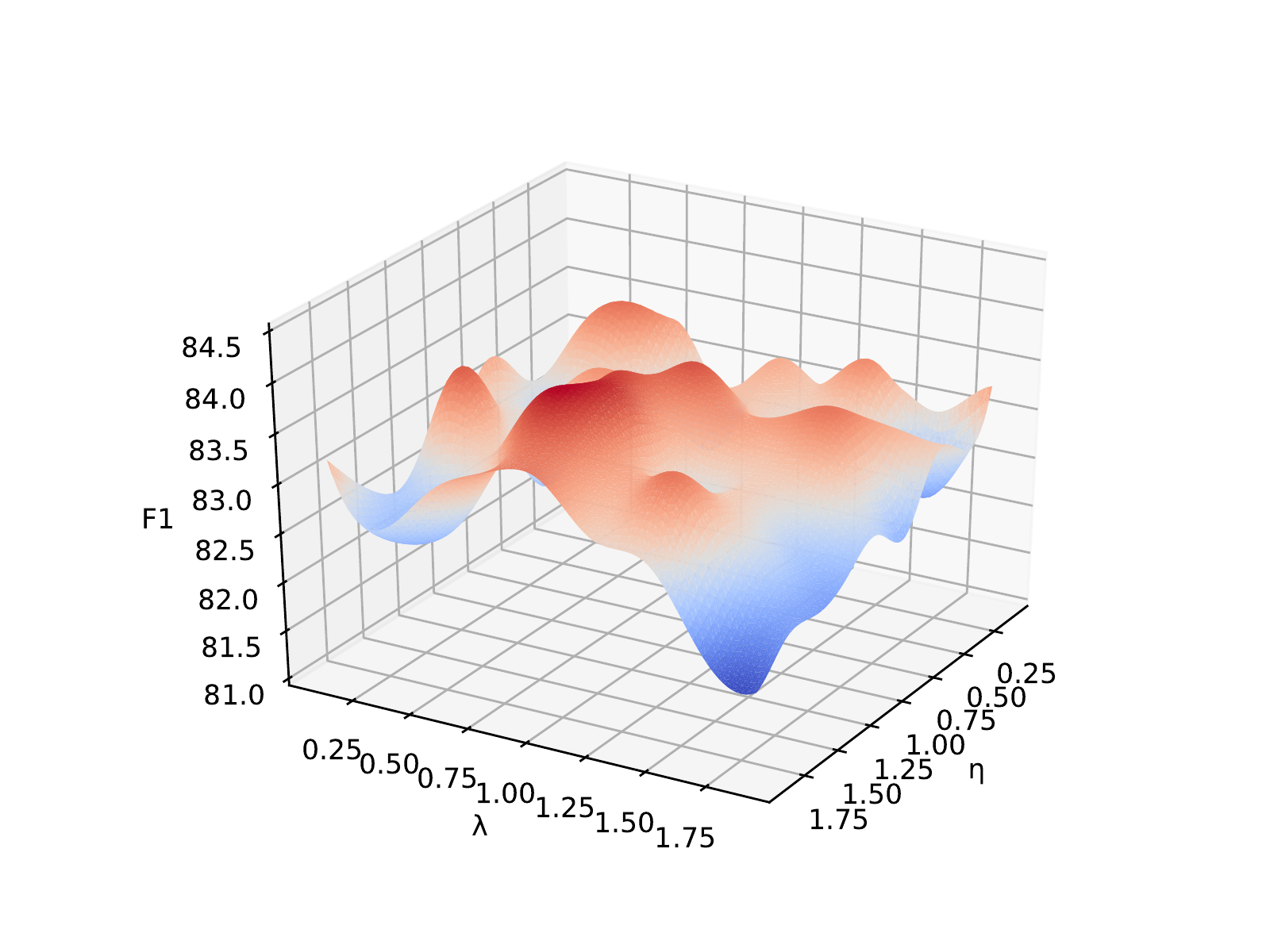}
\caption{Different performance with different values of $ \lambda$ and $\eta $. }
\label{fig:vis}
\end{figure}

In our experiments, we empirically set and fix the hyperparameters $\lambda$ to 0.1 and $\eta$ to 2 (see Eq.\ref{l}), and then tune the hyperparameters $\lambda$ and $\eta$ (see Eq.\ref{y}) using grid search. Fig.\ref{fig:vis} shows the different performance with different values of $ \lambda$ and $\eta $. We can see that our AnANet achieves the best performance when $\lambda=0.7$ and $\eta=1.3$.

\section{Conclusions}
In this paper, we define a new classification system (i.e. explicit relevant, implicit relevant, and irrelevant) for image-text correlation based on implicit association and explicit alignment. According to our proposed definition, we construct a two-stream method, AnANet, to model the essence of cross-modal correlation. Our model implicitly represents the global discrepancy and commonality between image and text and explicitly captures the cross-modal local relevance. The experimental results on our constructed new dataset demonstrate the effectiveness of our model. 

\bibliographystyle{IEEEtran}
\bibliography{ref}

\begin{thebibliography}{10}
\providecommand{\url}[1]{#1}
\csname url@samestyle\endcsname
\providecommand{\newblock}{\relax}
\providecommand{\bibinfo}[2]{#2}
\providecommand{\BIBentrySTDinterwordspacing}{\spaceskip=0pt\relax}
\providecommand{\BIBentryALTinterwordstretchfactor}{4}
\providecommand{\BIBentryALTinterwordspacing}{\spaceskip=\fontdimen2\font plus
\BIBentryALTinterwordstretchfactor\fontdimen3\font minus
  \fontdimen4\font\relax}
\providecommand{\BIBforeignlanguage}[2]{{%
\expandafter\ifx\csname l@#1\endcsname\relax
\typeout{** WARNING: IEEEtran.bst: No hyphenation pattern has been}%
\typeout{** loaded for the language `#1'. Using the pattern for}%
\typeout{** the default language instead.}%
\else
\language=\csname l@#1\endcsname
\fi
#2}}
\providecommand{\BIBdecl}{\relax}
\BIBdecl

\bibitem{ZEPPELZAUER201645}
M.~Zeppelzauer and D.~Schopfhauser, ``Multimodal classification of events in
  social media,'' \emph{Image and Vision Computing}, pp. 45--56, 2016.

\bibitem{sorodoc2017multimodal}
I.~Sorodoc, J.~H. Lau, N.~Aletras, and T.~Baldwin, ``Multimodal topic
  labelling,'' in \emph{Proceedings of the EACL}, 2017, pp. 701--706.

\bibitem{zadeh2017tensor}
A.~Zadeh, M.~Chen, S.~Poria, E.~Cambria, and L.-P. Morency, ``Tensor fusion
  network for multimodal sentiment analysis,'' in \emph{Proceedings of the
  EMNLP}, 2017, pp. 1103--1114.

\bibitem{zhang2018adaptive}
Q.~Zhang, J.~Fu, X.~Liu, and X.~Huang, ``Adaptive co-attention network for
  named entity recognition in tweets.'' in \emph{Proceedings of the AAAI},
  2018, pp. 5674--5681.

\bibitem{zhen2019deep}
L.~Zhen, P.~Hu, X.~Wang, and D.~Peng, ``Deep supervised cross-modal
  retrieval,'' in \emph{Proceedings of the CVPR}, 2019, pp. 10\,394--10\,403.

\bibitem{ma2019co}
R.~Ma, X.~Qiu, Q.~Zhang, X.~Hu, Y.-G. Jiang, and X.~Huang, ``Co-attention
  memory network for multimodal microblog's hashtag recommendation,''
  \emph{IEEE Transactions on Knowledge and Data Engineering}, 2019.

\bibitem{mccloud1998understanding}
S.~McCloud and A.~Manning, ``Understanding comics: The invisible art,''
  \emph{IEEE Transactions on Professional Communications}, vol.~41, no.~1, pp.
  66--69, 1998.

\bibitem{marsh2003taxonomy}
E.~E. Marsh and M.~D. White, ``A taxonomy of relationships between images and
  text,'' \emph{Journal of Documentation}, 2003.

\bibitem{martinec2005system}
R.~Martinec and A.~Salway, ``A system for image--text relations in new (and
  old) media,'' \emph{Visual communication}, vol.~4, no.~3, pp. 337--371, 2005.

\bibitem{chen2013understanding}
T.~Chen, D.~Lu, M.-Y. Kan, and P.~Cui, ``Understanding and classifying image
  tweets,'' in \emph{Proceedings of the ACM MM}, 2013, pp. 781--784.

\bibitem{wang2014bilateral}
Z.~Wang, P.~Cui, L.~Xie, W.~Zhu, Y.~Rui, and S.~Yang, ``Bilateral
  correspondence model for words-and-pictures association in multimedia-rich
  microblogs,'' \emph{ACM Transactions on Multimedia Computing, Communications,
  and Applications}, vol.~10, no.~4, pp. 1--21, 2014.

\bibitem{wu2014multimodal}
S.~Wu, ``A multimodal analysis of image-text relations in picture books,''
  \emph{Theory and Practice in Language Studies}, vol.~4, no.~7, p. 1415, 2014.

\bibitem{jas2015image}
M.~Jas and D.~Parikh, ``Image specificity,'' in \emph{Proceedings of the CVPR},
  2015, pp. 2727--2736.

\bibitem{chen2015velda}
T.~Chen, H.~M. SalahEldeen, X.~He, M.-Y. Kan, and D.~Lu, ``Velda: Relating an
  image tweet's text and images.'' in \emph{Proceedings of the AAAI}, 2015, pp.
  30--36.

\bibitem{henning2017}
C.~A. Henning and R.~Ewerth, ``Estimating the information gap between textual
  and visual representations,'' in \emph{Proceedings of the ICMR}, 2017, p.
  14–22.

\bibitem{otto2020characterization}
C.~Otto, M.~Springstein, A.~Anand, and R.~Ewerth, ``Characterization and
  classification of semantic image-text relations,'' \emph{International
  Journal of Multimedia Information Retrieval}, vol.~9, no.~1, pp. 31--45,
  2020.

\bibitem{zhang2018equal}
M.~Zhang, R.~Hwa, and A.~Kovashka, ``Equal but not the same: Understanding the
  implicit relationship between persuasive images and text,'' in
  \emph{Proceedings of the BMVC}, 2018, pp. 1--14.

\bibitem{kruk2019integrating}
J.~Kruk, J.~Lubin, K.~Sikka, X.~Lin, D.~Jurafsky, and A.~Divakaran,
  ``Integrating text and image: Determining multimodal document intent in
  instagram posts,'' in \emph{In Proceedings of the EMNLP-IJCNLP}, 2019, pp.
  4614--4624.

\bibitem{vempala2019categorizing}
A.~Vempala and D.~Preo{\c{t}}iuc-Pietro, ``Categorizing and inferring the
  relationship between the text and image of twitter posts,'' in
  \emph{Proceedings of the ACL}, 2019, pp. 2830--2840.

\bibitem{alikhani2019cite}
M.~Alikhani, S.~N. Chowdhury, G.~de~Melo, and M.~Stone, ``Cite: A corpus of
  image-text discourse relations,'' in \emph{Proceedings of the NAACL}, 2019,
  pp. 570--575.

\bibitem{alikhani2020cross}
M.~Alikhani, P.~Sharma, S.~Li, R.~Soricut, and M.~Stone, ``Cross-modal
  coherence modeling for caption generation,'' in \emph{Proceedings of the
  ACL}, 2020, pp. 6525--6535.

\bibitem{zhu2018msmo}
J.~Zhu, H.~Li, T.~Liu, Y.~Zhou, J.~Zhang, and C.~Zong, ``Msmo: Multimodal
  summarization with multimodal output,'' in \emph{Proceedings of the EMNLP},
  2018, pp. 4154--4164.

\bibitem{xu2018co}
N.~Xu, W.~Mao, and G.~Chen, ``A co-memory network for multimodal sentiment
  analysis,'' in \emph{Proceedings of the SIGIR}, 2018, pp. 929--932.

\bibitem{xu2020reasoning}
N.~Xu, Z.~Zeng, and W.~Mao, ``Reasoning with multimodal sarcastic tweets via
  modeling cross-modality contrast and semantic association,'' in
  \emph{Proceedings of the ACL}, 2020, pp. 3777--3786.

\bibitem{2015Faster}
S.~Ren, K.~He, R.~Girshick, and J.~Sun, ``Faster r-cnn: Towards real-time
  object detection with region proposal networks,'' \emph{IEEE Transactions on
  Pattern Analysis and Machine Intelligence}, vol.~39, no.~6, 2015.

\bibitem{2014GloVe}
J.~Pennington, R.~Socher, and C.~Manning, ``Glove: Global vectors for word
  representation,'' in \emph{Proceedings of the EMNLP}, 2014.

\bibitem{devlin-etal-2019-bert}
\BIBentryALTinterwordspacing
J.~Devlin, M.-W. Chang, K.~Lee, and K.~Toutanova, ``{BERT}: Pre-training of
  deep bidirectional transformers for language understanding,'' in
  \emph{Proceedings of the NAACL}, 2019, pp. 4171--4186. [Online]. Available:
  \url{https://www.aclweb.org/anthology/N19-1423}
\BIBentrySTDinterwordspacing

\bibitem{lee2018stacked}
K.-H. Lee, X.~Chen, G.~Hua, H.~Hu, and X.~He, ``Stacked cross attention for
  image-text matching,'' in \emph{Proceedings of the ECCV}, 2018.

\end{thebibliography}

\end{document}